\documentclass[preprint]{elsarticle}
\usepackage{amsmath,amsfonts}
\usepackage{array}
\usepackage{textcomp}
\usepackage{stfloats}
\usepackage{url}
\usepackage{verbatim}
\usepackage{graphicx}
\usepackage{multirow}
\usepackage{bbding}
\usepackage{gensymb}
\usepackage{booktabs}
\usepackage{subfigure}
\usepackage[ruled,linesnumbered]{algorithm2e}

\usepackage{color}

\usepackage{lineno,hyperref}

\begin{document}

\begin{frontmatter}

\title{Deep Image Clustering with Contrastive Learning and Multi-scale Graph Convolutional Networks}

\author[1]{Yuankun Xu}
\ead{ykxu@stu.scau.edu.cn}
\author[1,2]{Dong Huang \corref{cor1}}
\ead{huangdonghere@gmail.com}
\author[3,4]{Chang-Dong Wang}
\ead{changdongwang@hotmail.com}
\author[3]{Jian-Huang Lai}
\ead{stsljh@mail.sysu.edu.cn}
\address[1]{College of Mathematics and Informatics, South China Agricultural University, China}
\address[2]{Key Laboratory of Smart Agricultural Technology in Tropical South China, Ministry of Agriculture and Rural Affairs, China}
\address[3]{School of Computer Science and Engineering, Sun Yat-sen University, China}
\address[4]{Guangdong Provincial Key Laboratory of Intellectual Property and Big Data, China}
\cortext[cor1]{Corresponding author}

\begin{abstract}
Deep clustering has shown its promising capability in joint representation learning and clustering via deep neural networks. Despite the significant progress, the existing deep clustering works mostly utilize some distribution-based clustering loss, lacking the ability to unify representation learning and multi-scale structure learning. To address this, this paper presents a new deep clustering approach termed {\textbf{I}}mage {\textbf{c}}luster{\textbf{i}}ng with {\textbf{c}}ontrastive {\textbf{le}}arning and multi-scale \textbf{G}raph \textbf{C}onvolutional \textbf{N}etworks (IcicleGCN), which bridges the gap between convolutional neural network (CNN) and graph convolutional network (GCN) as well as the gap between contrastive learning and multi-scale structure learning for the deep clustering task. Our framework consists of four main modules, namely, the CNN-based backbone, the Instance Similarity Module (ISM), the Joint Cluster Structure Learning and Instance reconstruction Module (JC-SLIM), and the Multi-scale GCN module (M-GCN). Specifically, the backbone network with two weight-sharing views is utilized to learn the representations for the two augmented samples (from each image). The learned representations are then fed to ISM and JC-SLIM for joint instance-level and cluster-level contrastive learning, respectively, during which an auto-encoder in JC-SLIM is also pretrained to serve as a bridge to the M-GCN module. Further, to enforce multi-scale neighborhood structure learning, two streams of GCNs and the auto-encoder are simultaneously trained via (i) the layer-wise interaction with representation fusion and (ii) the joint self-adaptive learning. Experiments on multiple image datasets demonstrate the superior clustering performance of  IcicleGCN over the state-of-the-art. The code is available at \url{https://github.com/xuyuankun631/IcicleGCN}.

\end{abstract}


\end{frontmatter}

\section{Introduction}\label{sec:introduction}
Deep learning has achieved remarkable success in many supervised learning applications, which typically requires a considerable amount of training samples with true labels. {To alleviate the probably labor-intensive task of data annotation, the unsupervised learning techniques \cite{Fang2023a} have recently attracted increasing attention, among which the clustering analysis plays a fundamental role \cite{wang22_tcyb,lao23_tbd,Huang2023}.}

The traditional clustering methods \cite{wang22_tcyb,lao23_tbd,Huang2023} generally rely on hand-crafted features, which lack the ability of feature representation learning and may result in sub-optimal clustering performance for high-dimensional complex data. Recently some efforts have been made to incorporate the deep learning technique into the unsupervised clustering task. For example, Yang et al.\cite{yang2017towards} proposed the Deep Clustering Networks (DCN) method, where a reconstruction loss of the auto-encoder and a $K$-means based clustering loss are jointly optimized. Chang et al.\cite{chang2017deep} presented the Deep Adaptive image Clustering (DAC) method by formulating the clustering problem as a binary pairwise classification problem and enforcing the learned labels features to be one-hot vectors which can be used for image clustering. Xie et al.\cite{xie2016unsupervised} developed the Deep Embedded Cluster (DEC) method, which aims to map the learned features in the data space to a low-dimensional feature space with a Kullback-Leibler (KL) divergence based clustering loss. Yang et al. \cite{yang2016joint} devised the Joint Unsupervised LEarning (JULE) method, which iteratively updates a convolutional neural network (CNN) by performing the agglomerative clustering in the forward propagation and learning the feature representations in the backward propagation.

{Although these deep clustering works \cite{yang2017towards,chang2017deep,xie2016unsupervised,yang2016joint,wu2019deep,huang2020deep,Lu2022} have made significant progress, there are still some critical questions that remain to be addressed.} Especially, in this paper, we focus on the following three key questions.

\begin{enumerate}
	\item [Q1:] {Many of previous works tend to utilize some clustering loss (usually related to the label distributions) to guide the unsupervised training, which often overlook the sample-wise relationships in their learning process. With the contrastive learning recently showing its promising ability in self-supervised learning via positive and negative sample pairs \cite{Chen2020}, the first question arises as to \textbf{\textit{how to incorporate the contrastive learning into the deep clustering process for better representation learning and clustering}}}.
	\item [Q2:] {The conventional contrastive learning only considers the direct sample-wise relationships, e.g., the relationships between the positive pairs and the negative pairs. Regarding this, the second question arises as to \textbf{\textit{how to go beyond the direct sample-wise relationship to explore the rich information in neighborhood structures, or even enforce neighborhood structure learning for deep clustering}}}.
	\item [Q3:] Starting from the first two questions, the third question emerges as to \textbf{\textit{how to extend the neighborhood structure learning (in Q2) from single-scale to multi-scale, and jointly leverage multi-scale neighborhood structure learning and contrastive learning in a unified framework.}}
\end{enumerate}

More recently, several attempts have been carried out to address some of the above three questions. Li et al. \cite{li2021contrastive} proposed the Contrastive Clustering (CC) method to incorporate contrastive learning into deep clustering, which takes into account the sample-wise relationships (between positive sample-pairs and negative sample pairs) but still overlooks the sample-wise neighborhood structure.
To investigate the neighborhood structure,
van Gansbeke et al. \cite{van2020scan} presented a two-stage deep clustering method termed Semantic Clustering by Adopting Nearest neighbors (SCAN), where the first stage employs the contrastive learning  to learn the feature representation for constructing a $k$-nearest neighbor ($k$-NN) graph and the second stage aims to maximize the similarity between each sample and its $k$-NNs.
Zhong et al. \cite{zhong2021graph} designed the Graph Contrastive Clustering (GCC) method, which also utilizes a $k$-NN graph to provide more structure information for contrastive representation learning and clustering. Although SCAN \cite{van2020scan} and GCC \cite{zhong2021graph} have gone one step further to exploit the neighborhood structure information, yet they are still restricted to the static neighborhood connections and lack the ability to dynamically explore the higher-order connections via neighborhood structure learning.

In terms of neighborhood structure learning, the Graph Convolutional Network (GCN) provides an alternative and powerful tool \cite{kipf2016semi}. As an early attempt, Bo et al. \cite{bo2020structural} devised the Structural Deep Clustering Network (SDCN) method, which first pretrains an auto-encoder and then simultaneously trains the encoder and the GCN with a KL divergence loss.
With the incorporation of GCN \cite{bo2020structural}, the neighborhood structure learning can be enforced in SDCN. However, on the one hand, SDCN takes vectorized feature representations as input, which lacks convolutional layers to extract spatial information from complex image data. On the other hand, SDCN typically uses a single $k$-NN graph (which represents a specific scale of the neighborhood structure) as the initial graph for GCN, but ignores the possibilities of extending the neighborhood structure learning from a single scale to multiple scales, so as to explore more comprehensive structure information. It remains a challenging problem how to bridge \textit{the gap between CNN and GCN} as well as \textit{the gap between contrastive learning and neighborhood structure learning}, and further, how to go \textit{from single-scale to multi-scale} neighborhood structure learning in a unified deep image clustering framework.

\begin{figure}[!t]
	\begin{center}
		{\includegraphics[width=1\columnwidth]{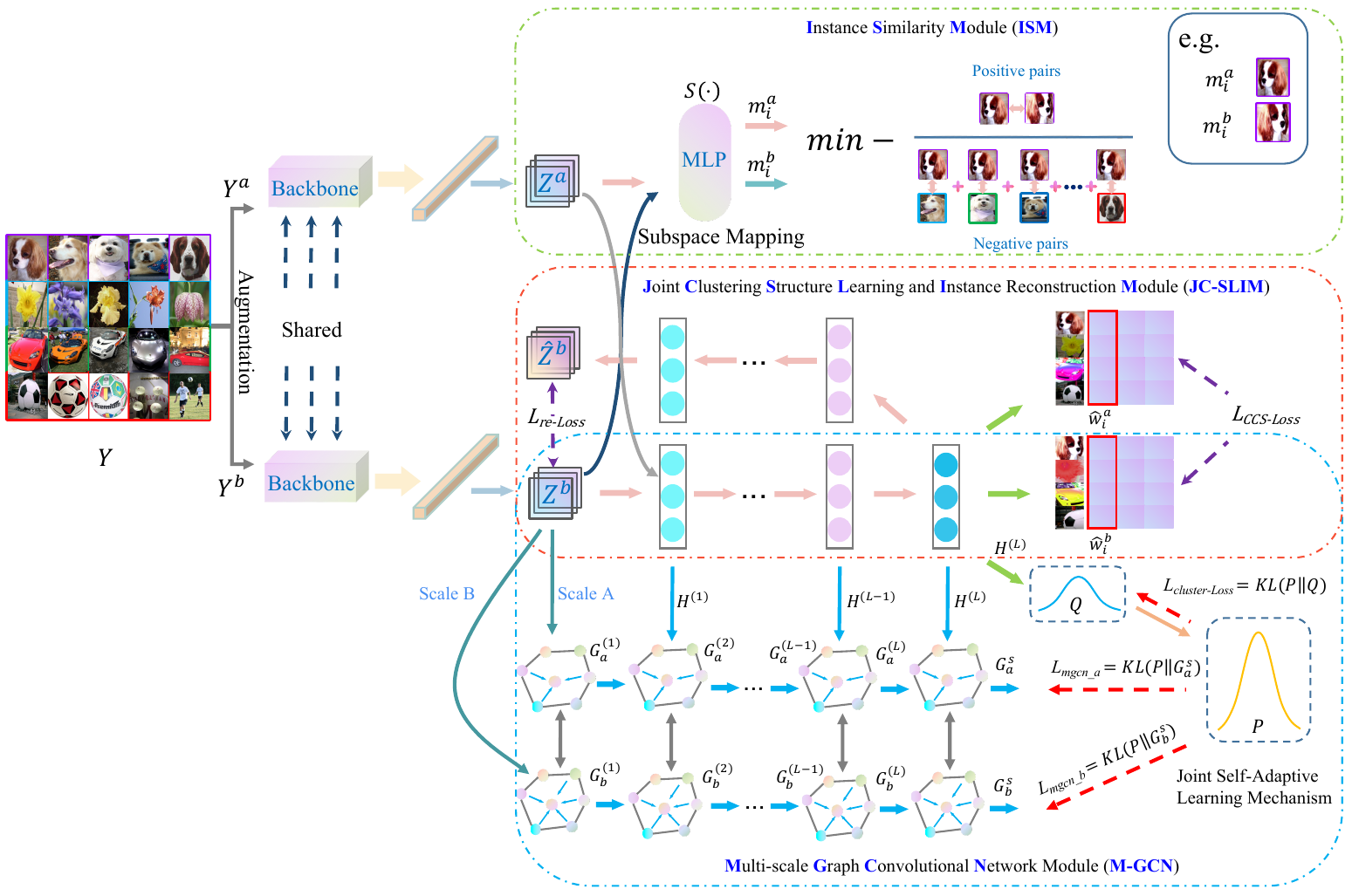}}
		\caption{The network architecture of \textbf{IcicleGCN} consists of four modules, i.e., the CNN-based \textbf{backbone} (which learns the representations for two augmentation views with shared weights), the \textbf{ISM} module (which enforces the instance-level contrastive learning), the \textbf{JC-SLIM} module (which enforces cluster-level contrastive learning and meanwhile pretrains an auto-encoder as a bridge to the neighborhood structure learning in the next module), and the \textbf{M-GCN} module (which jointly leverages multiple GCNs for multi-scale neighborhood structure learning).}
		\label{fig:mainstructure}
	\end{center}
\end{figure}

To address the above problem, in this paper, we present a novel deep clustering approach termed \textbf{I}mage \textbf{c}luster\textbf{i}ng with \textbf{c}ontrastive \textbf{le}arning and multi-scale \textbf{G}raph \textbf{C}onvolutional \textbf{N}etworks (\textbf{IcicleGCN}). The overall architecture of IcicleGCN consists of four modules, namely, the backbone network, the Instance Similarity Module (ISM), the Joint Cluster Structure Learning and Instance reconstruction Module (JC-SLIM), and the Multi-scale GCN module (M-GCN) (as shown in Fig.~\ref{fig:mainstructure}). Specifically, with two types of data augmentations randomly performed on each sample image, we utilize a weight-sharing CNN to learn the representations of the sample pairs, which are then fed to the later modules for contrastive learning and multi-scale neighborhood structure learning (via GCNs).
In ISM, we utilize a two-layer multilayer perceptron (MLP) with an instance-level contrastive loss is to maximize the similarity between positive pairs and minimize the similarity between negative pairs. In JC-SLIM, an auto-encoder is incorporated, which simultaneously exploits a cluster-level contrastive loss to learn the cluster structures and an instance reconstruction loss to pretrain the encoder that also serves as a bridge to the M-GCN module. {In M-GCN, two $k$-NN graphs are first built to represent two different scales of neighborhood structures, upon which two streams of GCNs and the auto-encoder (which is shared with the JC-SLIM module) are iteratively trained for the joint multi-scale neighborhood structure learning and clustering.  Extensive experiments are conducted on four image datasets, which demonstrate the superiority of the proposed IcicleGCN approach over the state-of-the-art deep clustering approaches.}

For clarity, the main contributions of this paper are summarized as follows:

\begin{enumerate}
	\item {This paper for the first time, to the best of our knowledge, enables multi-scale neighborhood structure learning for the image clustering task by taking advantage of multi-scale GCNs with joint self-adaptive learning.}
	\item 
	{This paper presents a novel deep image clustering approach termed IcicleGCN with instance-level contrastive learning, global cluster structure learning, and multi-scale neighborhood structure learning jointly enforced, which notably tackles the aforementioned three key questions (Q1, Q2, and Q3) in a unified framework.}
	\item {Extensive experiments have confirmed the superior clustering performance of our IcicleGCN approach on several challenging image datasets in comparison with the state-of-the-art deep clustering approaches.}
\end{enumerate}

The rest of the paper is organized as follows. Section~\ref{sec:related_work} reviews the related works on deep clustering and GCN. Section~\ref{sec:framework} describes the overall framework of IcicleGCN. Section~\ref{sec:experiment} reports the experimental results. Finally, Section~\ref{sec:conclusion} concludes the paper.

\section{Related Work}
\label{sec:related_work}

In this section, we review the related works on deep clustering and GCN in Sections~\ref{sec:related_work_DC} and \ref{sec:related_work_GCN}, respectively.

\subsection{Deep Clustering}
\label{sec:related_work_DC}
{In the past decade, the deep learning has shown its advantageous ability in learning discriminative features from complex data \cite{zhang2023applications,chen2022mcam,Ling2023,huang23tetci}. To exploit the representation learning ability of deep learning in the unsupervised scenarios, the deep clustering has recently drawn significant attention, which aims to learn high-quality feature representations for the clustering task without supervision information \cite{yang2017towards,chang2017deep,xie2016unsupervised,yang2016joint}}. A considerable number of deep clustering methods have been designed in recent years. For example,
Yang et al.\cite{yang2017towards} combined the auto-encoder network with the clustering task, where an auto-encoder is trained to reconstruct original features and a $K$-means clustering loss is incorporated to learn the cluster structure.
Chang et al.\cite{chang2017deep} mapped the clustering problem into a binary pairwise classification framework to judge whether each image pair belongs to the same cluster, and clustered the images by the local maximum response of the label features in the deep neural network.
Xie et al.\cite{xie2016unsupervised} presented the DEC method, which aims to map the learned features in the data space to a low-dimensional feature space, where the KL-divergence loss is used to iteratively optimize the cluster structure. Huang et al.\cite{huang2014deep} first utilized an auto-encoder to learn simplified representation from raw data, and then exploited a local preservation constraint to preserve the local structural properties of the data. Yang et al. \cite{yang2016joint} performed deep clustering by iteratively updating the CNN with the agglomerative clustering conducted in the forward propagation and the feature representations learned in the backward propagation. 

Besides these deep clustering methods that utilize some clustering loss in the deep neural network for cluster structure learning, another popular direction in recent years is to incorporate the contrastive learning paradigm \cite{li2021contrastive,van2020scan,Deng2023} for better representation learning and clustering. 
Specifically, Li et al. \cite{li2021contrastive} presented an end-to-end deep clustering method which performs the instance-level contrastive learning and the cluster-level contrastive learning at the same time.
Van Gansbeke et al. \cite{van2020scan} developed a two-stage method termed SCAN. In the first stage, it performs contrastive learning to learn feature representation for finding the nearest neighbors of each image. In the second stage, it obtains the clustering result via a semantic clustering loss  with the nearest neighbors of each image considered \cite{van2020scan}.

\subsection{Graph Convolutional Network}
\label{sec:related_work_GCN}
The concept of graph neural network (GNN) was first proposed by Gori et al.\cite{gori2005new} and further elaborated in GNN*\cite{scarselli2008graph}. Early works on GNN tend to use neural networks to transmit neighborhood information in an iterative way until reaching a stable fixed point.

The significant success of CNN in the field of computer vision has inspired some researchers to focus on convolutional operators for learning neighborhood information, which give rise to the GCN.
Bruna et al.\cite{bruna2013spectral} designed a variant of graph convolution based on spectral graph theory, which uses the spectral domain network to generalize convolutional networks through graph Fourier transform. Henaff et al. \cite{henaff2015deep} trained a graph convolutional layer that can perform forward and back propagation given a Fourier matrix, an interpolation kernel, and weights. Defferrard et al.\cite{defferrard2016convolutional} developed ChebNet to generalize CNNs to graph data thoroughly using spectral theoretical formulations, where the graph signal filtering and the graph coarsening are performed in the feature extraction stage. These spectral-based methods generally learn the entire graph structure at the same time. Since the processing of the entire graph structure is computationally expensive, these spectral-based methods are often difficult to generalize to large datasets.

Besides the spectral-based methods, there is an increasing number of spatial-based graph convolution methods in recent years. Monti et al. \cite{monti2017geometric} developed the MoNet framework to generalize traditional CNNs to non-Euclidean spaces. Niepert et al. \cite{niepert2016learning} proposed a general and efficient framework termed PATCHY-SAN for representation learning on arbitrary graph networks, which constructs locally connected neighborhoods in the graph and performs operations such as convolution on these neighborhoods. Gao et al. \cite{gao2018large} presented the Large-scale learnable Graph Convolutional Networks (LGCN) method, which creates the learnable graph convolutional layers and converts general graph data into grid-structure data with regular convolution operations. These methods directly perform convolution operations on the graph domain by aggregating the information of the neighboring nodes, while ignoring the representation information of the data itself. Shi et al. \cite{shi22_tip} exploited graph convolution operations on multi-scale prototype graphs for face recognition with image sets, which relies on the prior knowledge of image sets for generating prototypes and cannot perform sample-wise contrastive learning and unsupervised clustering.
Recently, to exploit the GCN for the unsupervised clustering task, Bo et al. \cite{bo2020structural} presented the SDCN method, which combines the features obtained from each layer of an encoder with the features learned by the GCN, and iteratively optimizes the network via a KL-divergence loss. Although SDCN considers the representation information of the data itself and the sample-wise structural information, yet it only takes the single-scale neighborhood information (with a single $k$-NN graph) as input, which overlooks the potential opportunities of jointly utilizing multi-scale neighborhood structures. Furthermore, SDCN relies on vectorized features, which restricts it application for complex images due to its lack of image-wise convolutional layers.

\section{Proposed Framework}
\label{sec:framework}

In this section, we describe the proposed IcicleGCN approach, which bridges the gap between CNN and GCN and also the gap between contrastive learning and multi-scale neighborhood structure learning. As illustrated in the Fig.~\ref{fig:mainstructure}, IcicleGCN consists of four modules, the CNN-based backbone, the ISM module, the JC-SLIM module, and the M-GCN module, which will be described in Sections~\ref{sec:backbone}, \ref{sec:ISM}, \ref{sec:JC-SLIM}, and \ref{sec:MGCN}, respectively.

\subsection{Backbone}
\label{sec:backbone}
{Our IcicleGCN approach utilizes a CNN (i.e., ResNet-34) as the backbone, which incorporates two augmentation views with shared weights to produce two augmented samples for each input image.  Specifically, given a mini-batch of $N$ images, denoted as $\{y_1, y_2, \cdots, y_N\}$, two types of augmentations are randomly selected for each image, leading to a total of $2\cdot N$ augmented samples for each mini-batch, which will then be fed to the following three modules (i.e., ISM, JC-SLIM, and M-GCN) for the instance-level contrastive learning, the joint cluster-level contrastive learning and auto-encoder training, and the multi-scale neighborhood structure learning, respectively.}

\subsection{ISM}
\label{sec:ISM}

The purpose of contrastive learning is to train the neural network by maximizing the similarity between the positive pairs while minimizing the similarity between the negative pairs \cite{Chen2020}. In this section, we describe the ISM module, which enforces the instance-level contrastive learning.

For an input image $y_{i}$, two augmented samples are generated, denoted as $y_{i}^{a}$ and $y_{i}^{b}$, respectively. Then the two augmented samples from the same input image are regarded as a positive sample pair, while the other $2\cdot (N-1)$ sample pairs are regarded as the negative pairs.
For an augmented sample $y_{i}^{u}$, its feature representation learned by the backbone is denoted as $z_{i}^{u}$ (for $u\in\{a,b\}$).

In the ISM module, a neural network with two fully-connected layers is leveraged to map the representations $z_i^{a}$ and $z_i^{b}$ to a low-dimensional space, denoted as $m_{i}^{a}=S\left(z_{i}^{a}\right)$ and $m_{i}^{b}=S\left(z_{i}^{b}\right)$. Thereby, the similarity between a pair of samples can be computed by the cosine similarity, that is

\begin{align}
	s\left(m_{i}^{u_{1}}, m_{j}^{u_{2}}\right)=&\frac{\left(m_{i}^{u_{1}}\right)\left(m_{j}^{u_{2}}\right)^{\top}}{\left\|m_{i}^{u_{1}}\right\|\left\|m_{j}^{u_{2}}\right\|}, \\
	\text{$u_{1},u_{2}\in\{a,b\}$}&,\text{$i,j\in[1,N]$}.\nonumber
\end{align}

To maximize the similarity between positive pairs and minimize the similarity between negative pairs,  the contrastive instance similarity loss for a sample $y_{i}^{a}$ is defined as \cite{Chen2020}

\begin{align}
	{\ell}_{i}^{a}=-\log \left(\frac{e^{s(m_{i}^{a}, m_{i}^{b}) / \tau_{I}}}{\sum_{j=1}^{N} (e^{s(m_{i}^{a}, m_{j}^{a}) / \tau_{I}}+e^{s(m_{i}^{a}, m_{j}^{b}) / \tau_{I}})}\right),
\end{align}
where $\tau_{I}$ is the temperature parameter. With the two augmented samples of each image considered, we have the contrastive instance similarity loss (for the $N$ images in a mini-batch) as

\begin{align}
	\label{eq: CIS-loss}
	\mathcal{L}_{CIS-Loss}=\frac{1}{2 N} \sum_{i=1}^{N}\left(\ell_{i}^{a}+\ell_{i}^{b}\right).
\end{align}

Thereby, with the instance-level contrastive learning enforced, we proceed to incorporate the cluster-level contrastive learning and the multi-scale neighborhood structure learning in the following.

\subsection{JC-SLIM}
\label{sec:JC-SLIM}

In this section, we describe the JC-SLIM module, whose technical role is two-fold. On the one hand, it enforces the cluster-level contrastive learning. On the other hand, it also serves as a bridge between the contrastive learning via CNN and the neighborhood structure learning via GCN by sharing the pretrained auto-encoder with the M-GCN module.

Specifically, an auto-encoder is utilized in the JC-SLIM module, where the encoder maps the representations learned by the backbone to a low-dimensional space for the cluster structure learning (i.e., the cluster-level contrastive learning) and meanwhile collaborates with the decoder for the instance reconstruction. In the following, the contrastive cluster structure loss and the instance reconstruction loss will be presented in Sections~\ref{sec:acc loss } and \ref{sec:reconstruction loss}, respectively.

\subsubsection{Contrastive Cluster Structure Loss }
\label{sec:acc loss }

Besides the contrastive learning at the instance-level, it is also expected to learn the cluster structures by considering the consistency of the two augmentation views. The intuition is to maximize the similarity between the cluster distributions of the two augmentation views, which gives rise to the cluster-level contrastive learning \cite{li2021contrastive}.

Given the feature representations (of an augmented pair $y_{i}^a$ and $y_{i}^b$) learned by the backbone, i.e.,  $z_{i}^{a}$ and $z_{i}^{b}$, we respectively map them to the $K$-dimensional representations $w^{a}_i$ and $w^{b}_i$ via the encoder (with softmax operation) of the auto-encoder. Here, the $K$-dimensional vector $w^{a}_i$ can be viewed as the probability of the sample $y_{i}^{a}$ belonging to each of the $K$ clusters.

By mapping the $N$ pairs of augmented samples to the $K$-dimensional space via the encoder, we can obtain the representation matrices for the two augmentation views, denoted as $W^{a}\in \mathbb{R}^{N \times K}$ and $W^{b}\in \mathbb{R}^{N \times K}$, respectively. Here, $w^{u}_i$ is the $i$-th row in $W^{u}$ (for $u\in \{a,b\}$). Let $\hat{w}_{i}^{u}$ denote the $i$-th column in $W^{u}$, which indicates the probability of the $N$ samples belonging to the $i$-th cluster. Specifically, $\hat{w}_{i}^{a}$ and $\hat{w}_{i}^{b}$ are regarded as a positive cluster pair, while the other $2\cdot (K-1)$ cluster pairs are regarded as the negative pairs.
Then the contrastive cluster structure loss for $\hat{w}_{i}^{a}$ is defined as
\begin{align}
	\hat{\ell}_{i}^{a}=-\log \left(\frac{e^{s(\hat{w}_{i}^{a}, \hat{w}_{i}^{b}) / \tau_{C}}}{\sum_{j=1}^{K} (e^{s(\hat{w}_{i}^{a}, \hat{w}_{j}^{a}) / \tau_{C}}+e^{s(\hat{w}_{i}^{a}, \hat{w}_{j}^{b}) / \tau_{C}})}\right),
\end{align}
where $\tau_{C}$ is the temperature parameter.
With two augmentation views $\hat{w}_{j}^{a}$ and $\hat{w}_{j}^{b}$ considered, the contrastive cluster structure loss for the $K$ clusters is defined as
\begin{align}
	\label{eq: CCS-loss}
	\mathcal{L}_{CCS-Loss}=\frac{1}{2 K} \sum_{i=1}^{K}\left(\hat{\ell}_{i}^{a}+\hat{\ell}_{i}^{b}\right)-H(W^{a})-H(W^{b}),
\end{align}
where $H(W^{u})=-\sum_{i=1}^{K}\left(P\left(\hat{w}_{i}^{u}\right) \log P\left(\hat{w}_{i}^{u}\right)\right)$ (for $u\in \{a,b\}$) is an entropy term that is incorporated to avoid the trivial solution of assigning all samples to a single cluster, and $P\left(\hat{w}_{i}^{u}\right)=\frac{1}{N} \sum_{j=1}^{N} w_{ji}^{u}$, where $w_{ji}^{u}$ is the $(j,i)$-th entry in $W^{u}$ (for $u\in \{a,b\}$).

\subsubsection{Instance Reconstruction Loss}
\label{sec:reconstruction loss}

In JC-SLIM, the encoder of the auto-encoder is exploited for both the contrastive cluster structure learning and the neighborhood structure learning (for the next module of M-GCN). Thus, besides the contrastive cluster structure loss, we also incorporate the instance reconstruction loss to aid the training of the encoder, that is
\begin{align}
	\label{eq: re-loss}
	\mathcal{L}_{r e -Loss}&=\frac{1}{2 N} \sum_{i=1}^{N}\sum_{u=\{a,b\}}\left\|\mathbf{z}_{i}^{u}-\hat{\mathbf{z}}_{i}^{u}\right\|_{2}^{2}\nonumber\\
	&=\frac{1}{2 N}\sum_{u=\{a,b\}}\|\mathbf{Z}^{u}-\hat{\mathbf{Z}^{u}}\|_{F}^{2},
\end{align}
where $\hat{\mathbf{z}}_{i}^{u}$ is the representation reconstructed by the decoder.
Before the neighborhood structure learning in the M-GCN module, we first jointly train the backbone, the ISM module, and the JC-SLIM module via the loss function $\mathcal{L}_{1}$, that is
\begin{align}
	\label{eq: L1-loss}
	\mathcal{L}_{1}=\mathcal{L}_{CIS-Loss}+ \mathcal{L}_{CCS-Loss}+ \mathcal{L}_{re-Loss}.
\end{align}
It is noteworthy that, empirically, a linear combination of the three losses without additional hyper-parameters to adjust their influences can already lead to quite robust performance.
By optimizing the loss function $\mathcal{L}_{1}$, the instance-level and cluster-level contrastive learning and the pretraining of the auto-encoder (for the next module) are simultaneously enforced.
In the next section, the auto-encoder and the multiple GCNs will collaboratively perform the multi-scale neighborhood structure learning for exploring more indepth and versatile sample-wise relationships for deep clustering.

\subsection{M-GCN}
\label{sec:MGCN}
In this section, we describe the M-GCN module in detail, which exploits the auto-encoder pretrained in the JC-SLIM module, takes advantage of the multiple GCNs for multi-scale neighborhood structure learning, and unifies the these components via joint self-adaptive learning.

\subsubsection{Multi-Scale Neighborhood Structures}
\label{sec:knn}

The graph structure is widely used to capture the sample-wise relationship in data, where a $k$-NN graph (with a fixed $k$) can reflect the sample-wise neighborhood structure of a specific scale \cite{van2020scan,zhong2021graph,bo2020structural}.  Due to variations of different datasets or even the variations of local structures within a given dataset, a proper scale (with an optimal $k$) is not easy to be determined for the $k$-NN graph construction in practice. Though some previous deep clustering methods attempt to explore the neighborhood structure \cite{van2020scan,zhong2021graph,bo2020structural}, yet they generally rely on a single $k$-NN graph and cannot go beyond the single-scale neighborhood structure to explore the diversity in multiple scales. In light of this,  in the M-GCN module of our IcicleGCN framework, multiple scales of neighborhood structures are jointly investigated via multiple GCNs.

Specifically, in the M-GCN module, two $k$-NN graphs are initialized to capture the information of different scales of neighborhood structures, which are then jointly propagated and refined by two streams of GCNs. Note that M-GCN can also be extended to three or more streams of GCNs. Yet we empirically find that two scales of neighborhood structure have brought in sufficient improvement and thus we formulate the M-GCN module with two-stream GCNs (coupled with the auto-encoder from the JC-SLIM module).

With the backbone and the auto-encoder trained via the loss function $\mathcal{L}_{1}$ with the instance-level and cluster-level contrastive learning enforced, this section focuses on the neighborhood structure learning and does not involve the contrastive learning, which means that the augmentation pairs are no longer required here. Therefore, for each input image   $y_i$, it is augmented once, and then the backbone obtains its representation $z^{i}$. The representations of $N$ samples can be stacked as a feature matrix $Z^{b} \in \mathbb{R}^{N \times d}$, where $d$ is the output dimension of the backbone.
To initialize the multi-scale neighborhood structures, we need to construct $k$-NN graphs of different scales from the feature matrix, and then aggregate the multi-scale neighborhood information from different $k$-NN graphs via the propagation of GCNs. Here, to define the similarity between two samples, say, $z_i$ and $z_j$, the heat kernel \cite{bo2020structural} is utilized, that is
\begin{align}
	\mathrm{S}_{i j}=e^{-\frac{\left\|\mathrm{z}_{i}-\mathrm{z}_{j}\right\|^{2}}{t}},
\end{align}
where $t$ is the time parameter in the heat kernel. In IcicleGCN, we use $t=1$ for all experiments.
Further, we construct two different $k$-NN graphs to reflect two different scales of neighborhood structures, whose numbers of nearest neighhors are set to $k_a$ and $k_b$ (with $k_a\neq k_b$), respectively. Let $A_a$ and $A_b$ denote the adjacency matrices of the two constructed $k$-NN graphs. In the next section, we will utilize these two $k$-NN graphs and the feature embedding learned by the backbone as the input to the trident network with GCNs.

\subsubsection{Trident Network Architecture with GCNs}
\label{sec:trident network}

In the M-GCN module, three sources of input are taken, namely, the feature embedding which is learned by the backbone and fed to the auto-encoder, and the two scales of $k$-NN graphs which are fed to the two streams of GCNs. Thereby, three streams of networks (including an auto-encoder and two GCNs) are exploited in M-GCN, which is called a trident network architecture for our multi-scale neighborhood structure learning in the proposed IcicleGCN framework (as shown in Fig.~\ref{fig:mainstructure}).

Formally, in terms of the auto-encoder, let $L$ be the number of layers and $\mathbf{H}^{(\ell)}$ be the learned feature embedding at the $l$-th layer of its encoder. Thus the feedforward propagation of the encoder can be represented as

\begin{align}
	\mathbf{H}^{(\ell)}=\Phi \left(\mathbb{W}_{e}^{(\ell)} \mathbf{H}^{(\ell-1)}+ \beta_{e}^{(\ell)}\right),
\end{align}
where $\Phi$ is the activation function in the encoder, such as ReLU, and $\mathbb{W}_{e}^{(\ell)}$ and $\beta_{e}^{(\ell)}$ are the weight matrix and the bias of the encoder at the $\ell$-th layer, respectively. In particular, we use the feature matrix $\mathbf{Z}^{b}$ obtained from the backbone as the input $\mathbf{H}^{(0)}$.

Similarly, with the decoding layers corresponding to the encoding layers, each decoding layer reconstructs the input data as follows:
\begin{align}
	\mathbf{H}_{d}^{(\ell)}=\Phi \left(\mathbb{W}_{d}^{(\ell)} \mathbf{H}_{d}^{(\ell-1)}+ \beta_{d}^{(\ell)}\right),
\end{align}
where $\mathbb{W}_{d}^{(\ell)}$ and $\beta_{d}^{(\ell)}$ represent the weight matrix and the bias of the decoder at the $\ell$-th layer, respectively.

For convenience, we denote the output of the last decoder layer (i.e., $\mathbf{H}_{d}^{(L)}$) as $\hat{\mathbf{Z}^{b}}$. Then the reconstruction loss of the auto-encoder is represented as
\begin{align}
	\mathcal{L}_{mgcn\_re -Loss}=&\frac{1}{N}\|\mathbf{H}^{(0)}-\mathbf{H}_{d}^{(L)}\|_{F}^{2}\nonumber\\
	\label{eq:mgcn_re_loss}
	=&\frac{1}{N}\|\mathbf{Z}^{b}-\hat{\mathbf{Z}^{b}}\|_{F}^{2}.
\end{align}

Note that the loss~(\ref{eq:mgcn_re_loss}) is similar to but in fact different from the loss~(\ref{eq: re-loss}). The loss~(\ref{eq: re-loss}) is further used to collaborate with the contrastive loss in the unified loss~(\ref{eq: L1-loss}) of the ISM and JC-SLIM modules, where both augmentation views should be incorporated for contrastive learning and thus the reconstruction loss of both augmented samples (of each input sample) should also be considered. Since the contrastive learning is not involved in the M-GCN module, the necessity to consider both augmentation views no longer holds, which gives rise to the reconstruction loss~(\ref{eq:mgcn_re_loss}).

Further, two streams of GCNs are incorporated in the M-GCN module to \emph{interact} with the encoder of the auto-encoder. Before introducing the interaction (in the layer-wise update and the joint self-adaptive learning), we first describe the GCN and its learning process.

Each sample is regarded as a vertex in the graph.  Let $V$ denote the set of all the sample vertices and $A$ denote the adjacency matrix. Then the $i$-th sample $V_{i}$ aggregates the neighborhood information as
\begin{align}
	\operatorname{Aggregate}(V) &=\widetilde{D}^{-\frac{1}{2}} \widetilde{A}\widetilde{D}^{-\frac{1}{2}} V
\end{align}
\begin{align}
	\left(\widetilde{D}^{-\frac{1}{2}} \widetilde{A} \widetilde{D}^{-\frac{1}{2}} V\right)_{i} &=\left(\widetilde{D}^{-\frac{1}{2}} \widetilde{A}\right)_{i} \widetilde{D}^{-\frac{1}{2}} V \nonumber \\
	&=\left(\sum_{k} \widetilde{D}_{i k}^{-\frac{1}{2}} \widetilde{A}_{i}\right) \widetilde{D}^{-\frac{1}{2}} V \nonumber \\
	&=\widetilde{D}_{i i}^{-\frac{1}{2}} \sum_{j} \widetilde{A}_{i j} \sum_{k} \widetilde{D}_{j k}^{-\frac{1}{2}} V_{j} \nonumber \\
	&=\widetilde{D}_{i i}^{-\frac{1}{2}} \sum_{j} \widetilde{A}_{i j} \widetilde{D}_{j j}^{-\frac{1}{2}} V_{j} \nonumber \\
	&=\sum_{j} \frac{1}{\sqrt{\widetilde{D}_{i i} \widetilde{D}_{j j}}} \widetilde{A}_{i j} V_{j},
\end{align}
where $\widetilde{A} = A + I$, $I$ is the identity diagonal matrix, and $\widetilde{{D}}$ is the degree matrix with $\widetilde{{D}}_{i i}=\sum_{j} \widetilde{{A}}_{{i j}}$. The purpose of $\widetilde{A}$ is to obtain the information of the node itself in the adjacency matrix $A$.

Then we proceed to describe the learning process of the two-stream GCNs.  Let $\mathbf{G}^{(\ell)}_a$ and $\mathbf{G}^{(\ell)}_b$ denote the representations respectively learned by the \textit{first} and the \textit{second} GCNs at the $\ell$-th layer, where the graph convolution operation can be conducted as
\begin{align}
	\label{eq:multiwayequation}
	\mathbf{G}^{(\ell)}_u=\Phi\left(\widetilde{{D}}^{-\frac{1}{2}} \widetilde{{A}}_u \widetilde{{D}}^{-\frac{1}{2}}\mathbf{G}^{(\ell-1)}_u \mathbb{W}^{(\ell-1)}_u\right),
	\text{~for~} u\in\{a,b\},
\end{align}
where $\mathbb{W}^{\ell-1}_u$ (for $u\in\{a,b\})$ is the weight matrix of the ${(\ell-1)}$-th layer of the corresponding GCN. Under the guidance of the symmetric normalized adjacency matrix $\widetilde{{D}}^{-\frac{1}{2}} \widetilde{{A}}_u \widetilde{{D}}^{-\frac{1}{2}}$, we use the representation $\mathbf{G}^{(\ell-1)}_u$ of the ${(\ell-1)}$-th layer to generate the representation $\mathbf{G}^{(\ell)}_u$ of the $\ell$-th layer through the graph convolution operation~(\ref{eq:multiwayequation}).

Besides the two streams of GCNs, the representation learned by the $l$-th layer of the encoder (of the auto-encoder) is denoted as $\mathbf{H}^{(\ell)}$. Then we update each of the two-stream GCNs as follows:
\begin{align}
	\widetilde{\mathbf{G}}_{a}^{(\ell-1)}=\sigma \mathbf{G}_{a}^{(\ell-1)} + \gamma \mathbf{G}_{b}^{(\ell-1)} +(1-\sigma-\gamma) \mathbf{H}^{(\ell-1)},\\
	\widetilde{\mathbf{G}}_{b}^{(\ell-1)}=\sigma \mathbf{G}_{b}^{(\ell-1)} + \gamma \mathbf{G}_{a}^{(\ell-1)} +(1-\sigma-\gamma) \mathbf{H}^{(\ell-1)},
\end{align}
where $\sigma$ and $\gamma$ are two balance coefficients. In this work, we set $\sigma$ to 0.4 and $\gamma$ to 0.2 on all experiments. 
Through this layer-by-layer connection, the representation learning processes of the encoder and the two-stream GCNs are jointly leveraged.

Thereafter, we feed $\widetilde{\mathbf{G}}_{u}^{(\ell-1)}(u\in\{a,b\})$ into the $\ell$-th layer to generate the representation $\widetilde{\mathbf{G}}_{u}^{(\ell)}$ as

\begin{align}
	\widetilde{\mathbf{G}}_{u}^{(\ell)}=\Phi\left(\widetilde{{D}}^{-\frac{1}{2}} \widetilde{{A}}_u \widetilde{{D}}^{-\frac{1}{2}} \widetilde{\mathbf{G}}_{u}^{(\ell-1)} \mathbb{W}^{(\ell-1)}_u\right).
\end{align}

The GCN can aggregate the neighborhood information of each node in its first layer, which is called the first-order neighborhood information. And the neighbors are also aggregating their own neighborhood information. Therefore, In the second layer, when the same node aggregates its neighborhood information again, the neighborhood information of its neighbors can be aggregated, which is called the second-order neighborhood information. Theoretically, with the number of layers increased to sufficiently large, a node can aggregate the information of all nodes in the $k$-NN graph. However, in practical applications, the number of GCN layers generally will not exceed 4 or 5. In fact, as the number of layers increases to a large number, the aggregation of such a large amount of information will make the nodes in the graph undistinguishable, which is called the over-smoothing and may degrade the model performance.

In our M-GCN module, we use five layers of GCN, in which the last layer is associated with a softmax operation.
Note that the feature representation $\mathbf{Z}^{b}$ obtained from the backbone is used as the original input of the node features (to the first GCN layer), that is
\begin{align}
	\mathbf{G}^{(1)}_u=\Phi\left(\widetilde{{D}}^{-\frac{1}{2}} \widetilde{{A}}_u \widetilde{{D}}^{-\frac{1}{2}} \mathbf{Z}^{b} \mathbb{W}^{(0)}_u\right),
\end{align}

The last layer of the GCN is associated with a softmax operation, that is
\begin{align}
	\label{eq:GCN-loss}
	\mathbf{G}^{s}_u=\operatorname{softmax}\left( \mathbf{G}_u^{(L)} \right).
\end{align}

Thus, with the dimension of the softmax layer set to the number of clusters $K$, we obtain two probability matrices $\mathbf{G}^{s}_u$ (with $u\in\{a,b\}$) for the cluster assignments,
where the $(i,j)$-th entry of $\mathbf{G}^{s}_u$ represents the probability that the sample $i$ should be assigned to cluster $j$. Therefore, $\mathbf{G}^{s}_{a}$ and $\mathbf{G}^{s}_{b}$ can be regarded as two probability distributions, which will be exploited in the joint self-adaptive learning process.

\subsubsection{Joint Self-Adaptive Learning}
\label{sec: self-adaptive}

With the auto-encoder and the two streams of GCNs connected via the layer-wise updating, the consistency of their final-layer representations will be our next focus.
In this section, we design a joint self-adaptive learning mechanism to simultaneously guide the learning of the three streams of networks in M-GCN.

First, we use the Student's t-distribution \cite{van2008visualizing} to convert the representations obtained by the auto-encoder into a probability distribution. That is, the probability that the $i$-th sample is assigned to the $j$-th cluster can be obtained as follows:
\begin{align}
	\label{eq:Q-loss}
	q_{i j}=\frac{\left(1+\left\|\mathbf{h}_{i}-\boldsymbol{\mu}_{j}\right\|^{2} / t\right)^{-\frac{t+1}{2}}}{\sum_{j^{\prime}}\left(1+\left\|\mathbf{h}_{i}-\boldsymbol{\mu}_{j^{\prime}}\right\|^{2} / t\right)^{-\frac{t+1}{2}}},
\end{align}
where $t$ is the degree of freedom of the Student's t-distribution in the probability distribution function, and $\boldsymbol{\mu}_{j}$ is the $j$-th cluster center initialized by $K$-means  through the pre-trained representations of the auto-encoder. In the encoder part, we obtain the data representation matrix $\mathbf{H}^{L}$, where the $i$-th row $\mathbf{h}_{i}$ is the representation of the $i$-th sample. Here, the probability of assigning sample $i$ to cluster $j$ (i.e., $q_{ij}$) is obtained by computing the similarity between $\mathbf{h}_{i}$ and $\boldsymbol{\mu}_{j}$. By considering the probability distribution that the samples assigned to different clusters, we can denote the probability distribution matrix as $Q$, where $q_{ij}$ is its $(i,j)$-th entry.

To make the representations obtained from the auto-encoder closer to the center of the corresponding cluster, we proceed to compute a target distribution $P$ from $Q$, whose $(i,j)$-th entry $p_{i j}$ is defined as
\begin{align}
	\label{eq:P-loss}
	p_{i j}=&\frac{q_{i j}^{2} / f_{j}}{\sum_{j^{\prime}} q_{i j^{\prime}}^{2} / f_{j^{\prime}}},
\end{align}
with
\begin{align}
	f_{j}=&\sum_{i} q_{i j}.
\end{align}

Since $q_{ij}$ is a soft assignment probability, $f_{j}$ can thus be regarded as the soft cluster assignment frequency. In turn, we can use the target distribution $P$ to supervise the learning of the distribution $Q$:
\begin{align}
	\mathcal{L}_{cluster-Loss}=K L(P \| Q)=\sum_{i} \sum_{j} p_{i j} \log \frac{p_{i j}}{q_{i j}}.
\end{align}
By this formulation, we then perform self-adaptive learning of the $Q$ distribution by minimizing the KL-divergence between the $P$ and $Q$ distributions. 

Further, by regarding the representations $\mathbf{G}^{s}_{a}$ and $\mathbf{G}^{s}_{b}$ respectively learned by the two GCNs as probability distributions, we can also use the probability distribution $P$ to supervise the update and learning of the two-stream GCNs by minimizing their KL-divergence losses as follows:
\begin{align}
	\mathcal{L}_{mgcn_{-}a}=K L(P \|\mathbf{G}^{s}_{a})=\sum_{i} \sum_{j} p_{i j} \log \frac{p_{i j}}{g^{s}_{a,ij}},
\end{align}

\begin{align}
	\mathcal{L}_{mgcn_{-}b}=K L(P \|\mathbf{G}^{s}_{b})=\sum_{i} \sum_{j} p_{i j} \log \frac{p_{i j}}{g^{s}_{b,ij}},
\end{align}
where $g^{s}_{a,ij}$ and $g^{s}_{b,ij}$ denote the $(i,j)$-th entries of $\mathbf{G}^{s}_{a}$ and $\mathbf{G}^{s}_{b}$, respectively. For the joint self-adaptive learning of the auto-encoder and the two-stream GCNs, we define the overall loss function $\mathcal{L}_{2}$ of the M-GCN module as
\begin{align}
	\label{eq:L2-loss}
	\mathcal{L}_{2}=&\mathcal{L}_{mgcn\_re -Loss}+\alpha \mathcal{L}_{cluster-Loss} \nonumber\\
	&+\beta \mathcal{L}_{mgcn_{-}a}+\eta \mathcal{L}_{mgcn_{-}b},
\end{align}
where $\alpha$, $\beta$, and $\eta$ are hyper-parameters to  balance the influences of different terms. 
Through this formulation, the information of the data itself and the multi-scale neighborhood structure information can be adaptively aggregated, and the final clustering can be obtained from the probability distribution learned by the GCNs. {For clarity, the overall process of our IcicleGCN approach is described in Algorithm~1.}

\begin{algorithm}[!h]
	\label{alg}
	\caption{Training algorithm for IcicleGCN.}
	\LinesNumbered
	\KwIn{Image dataset $\mathbf{Y}$, the training epochs ${E}$, the batch size ${N}$, the number of iterations in M-GCN ${N}_{it}$, the temperature parameters $\tau_I$ and $\tau_C$, and the number of clusters $K$.}
	\KwOut{The clustering result with $K$ clusters.}
	\For{each epoch}{
		$\mathbf{Step 1:}$ Randomly select a mini-batch of ${N}$ images from the image dataset\;
		$\mathbf{Step 2:}$ Perform two random data augmentations on each image in the mini-batch\;
		$\mathbf{Step 3:}$ Calculate the contrastive instance similarity loss by Eq. (\ref{eq: CIS-loss})\;
		$\mathbf{Step 4:}$ Calculate the contrastive cluster structure loss by Eq. (\ref{eq: CCS-loss})\;
		$\mathbf{Step 5:}$ Calculate the instance reconstruction loss by Eq. (\ref{eq: re-loss})\;
		$\mathbf{Step 6:}$ Jointly update the backbone, the ISM, and the JC-SLIM by minimizing $\mathcal{L}_{1}$ in Eq. (\ref{eq: L1-loss})\;
	}
	\For{each iteration in ${N}_{it}$}{
		$\mathbf{Step 7:}$ Calculate the probability distribution of the two-stream GCN networks by Eq. (\ref{eq:GCN-loss})\;
		$\mathbf{Step 8:}$ Calculate the $Q$ distribution by Eq. (\ref{eq:Q-loss})\;
		$\mathbf{Step 9:}$ Calculate the $P$ distribution by Eq. (\ref{eq:P-loss})\;
		$\mathbf{Step 10:}$ Update the GCNs and the auto-encoder by minimizing $\mathcal{L}_{2}$ in Eq. (\ref{eq:L2-loss}) \;
	}
	Obtain the final clustering from the probability distribution learned by the M-GCN module\;
\end{algorithm}

\subsection{Implementation Details}
\label{sec:BGCG}

Our IcicleGCN approach consists of four modules, including the backbone, the ISM module, the JC-SLIM module, and the M-GCN module. Specifically, we adopt the ResNet-34 as the backbone network, and resize the images in the dataset to the size of $224\times 224\times 3$. Note that there is an overlapping (or sharing) component between the JC-SLIM module and the M-GCN module, that is, the auto-encoder, which is pretrained in JC-SLIM and further participates in the updating process of the multi-scale GCNs. As for the network training, we first use the loss function $\mathcal{L}_{1}$ to train the backbone, the ISM, and the JC-SLIM, and then use the loss function $\mathcal{L}_{2}$ to train the M-GCN. The dimension of the auto-encoder is set to input-500-500-2000-$K$, where $K$ is the number of clusters.
The batch size is set to 128. We use the Adam optimizer with a learning rate of 1e-4. The parameters $\sigma$ and $\gamma$ are set to 0.4 and 0.2, respectively. The parameters $\alpha$, $\beta$ and $\eta$ in the loss function $\mathcal{L}_{2}$ are tuned in the range of $\{$0.01, 0.05, 0.1$\}$. To construct the two different $k$-NN graphs, the numbers of nearest neighbors are set of 1 and 10, respectively. That is, a $1$-NN graph and a $10$-NN graph are initialized for M-GCN. All experiments are carried out on a machine with an Nvidia GeForce RTX 3090 GPU and a CPU with 12 cores and 2.6GHz.

\begin{figure}[!t]
	\centering
	\subfigure[CIFAR-10]{
		\includegraphics[width=0.235\textwidth]{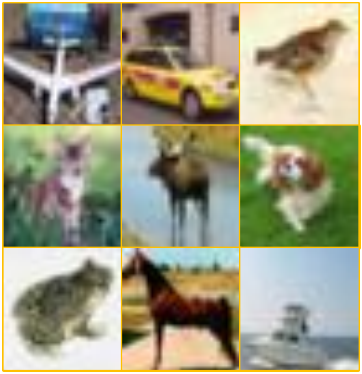}}
	\subfigure[CIFAR-100]{
		\includegraphics[width=0.235\textwidth]{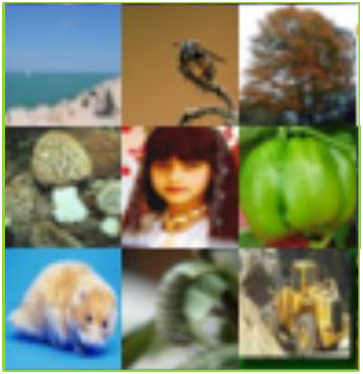}}
	\subfigure[ImageNet-10]{
		\includegraphics[width=0.235\textwidth]{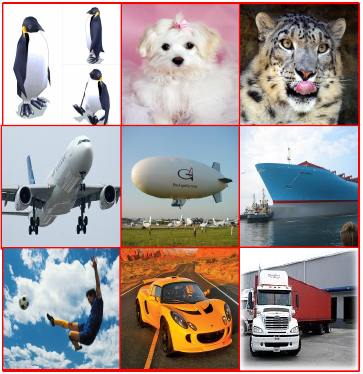}}
	\subfigure[ImageNet-Dogs]{
		\includegraphics[width=0.235\textwidth]{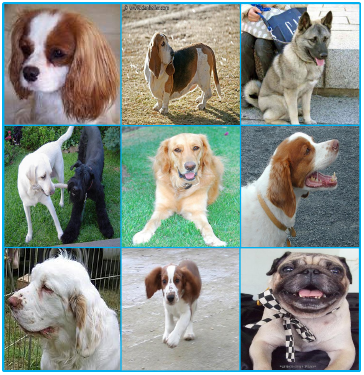}}
	\caption{Some examples of the four image datasets.}\vskip 0.15 in
	\label{fig:dataset_samples}
\end{figure}

\section{Experiments}
\label{sec:experiment}

In this section, we experimentally compare the proposed IcicleGCN algorithm against several deep and non-deep clustering algorithms on multiple image datasets.

\subsection{Datasets and Evaluation Metrics}

We conduct experiments on four image datasets. Some sample images of these datasets are shown in Fig.~\ref{fig:dataset_samples}. The details of the four datasets are given below.
\begin{itemize}
	\item \textbf{CIFAR-10 \cite{krizhevsky2009learning}:} The CIFAR-10 dataset consists of 60,000 color images of size 32$\times$32$\times$3. These images are divided into 10 classes, including airplanes, cars, birds, cats, trucks, etc.
	\item \textbf{CIFAR-100 \cite{krizhevsky2009learning}:} The CIFAR-100 datasets consists of 60,000 color images of size 32$\times$32$\times$3. Unlike CIFAR-10, each image in CIFAR-100 has an ``elaborate" class label (corresponding to a total of 100 classes) and a ``coarse" class label (corresponding to a total of 20 super-classes). In this paper, the 20 super-classes are taken as the ground-truth.
	\item \textbf{ImageNet-10 \cite{chang2017deep}:} The ImageNet-10 dataset is a subset of the ImageNet dataset, containing a total of 13,000 color images with 10 classes. The size of each image is 96$\times$96$\times$3.
	\item \textbf{ImageNet-Dogs \cite{chang2017deep}:} The ImageNet-Dogs dataset is also a subset of the ImageNet dataset, containing 19,500 images with 15 dog categories. The size of each image is 96$\times$96$\times$3.
\end{itemize}

To quantitatively compare the clustering results by different clustering methods, we adopt three widely-used evaluation metrics, namely, Accuracy (ACC) \cite{Fang2023}, Normalized Mutual Information (NMI) \cite{Liang2022}, and Adjusted Rand Index (ARI) \cite{Huang2021}. Notice that larger values of the three metrics indicate better clustering performance.

\begin{table}[!t]
	\caption{The clustering performance w.r.t. \textbf{ACC(\%)} by different clustering methods on the benchmark datasets}\vskip 0.05 in
\label{table:ACC}
\centering
\setlength{\tabcolsep}{1.45mm}{
	\begin{tabular}{|c|c|c|c|c|}
		\hline
		Dataset                                & CIFAR-10 & CIFAR-100 & ImageNet-10 & ImageNet-Dogs \\ \hline \hline
		K-means \cite{macqueen1967some}        & 22.9     & 13.0      & 24.1        & 10.5          \\ \hline
		SC      \cite{zelnik2004self}          & 24.7     & 13.6      & 27.4        & 11.1          \\ \hline
		AC      \cite{gowda1978agglomerative}  & 22.8     & 13.8      & 24.2        & 13.9          \\ \hline
		NMF     \cite{2009Locality}            & 19.0     & 11.8      & 23.0        & 11.8          \\ \hline
		AE      \cite{bengio2006greedy}        & 31.4     & 16.5      & 31.7        & 18.5          \\ \hline
		DAE     \cite{vincent2010stacked}      & 29.7     & 15.1      & 30.4        & 19.0          \\ \hline
		DCGAN   \cite{radford2015unsupervised} & 31.5     & 15.3      & 34.6        & 17.4          \\ \hline
		DeCNN   \cite{zeiler2010deconvolutional} & 28.2     & 13.3      & 31.3        & 17.5          \\ \hline
		VAE     \cite{kingma2013auto}          & 29.1     & 15.2      & 33.4        & 17.9          \\ \hline
		JULE    \cite{yang2016joint}           & 27.2     & 13.7      & 30.0        & 13.8          \\ \hline
		DEC     \cite{xie2016unsupervised}     & 30.1     & 18.5      & 38.1        & 19.5          \\ \hline
		DAC     \cite{chang2017deep}           & 52.2     & 23.8      & 52.7        & 27.5          \\ \hline
		DDC     \cite{chang2019deep}           & 52.4     & -         & 57.7        & -             \\ \hline
		DCCM    \cite{wu2019deep}              & 62.3     & 32.7      & 71.0        & 38.3          \\ \hline
		IIC     \cite{ji2019invariant}         & 61.7     & 25.7      & -           & -             \\ \hline
		GATCluster \cite{niu2020gatcluster}    & 62.3     & 32.7      & 73.9        & 32.2          \\ \hline
		PICA    \cite{huang2020deep}           & 69.6     & 33.7      & 87.0         & 35.2          \\ \hline
		DRC     \cite{zhong2020deep}           & 72.7     & 36.7      & 88.4        & 38.9          \\ \hline
		CC      \cite{li2021contrastive}       & 79.0     & 42.9      & 89.5        & 34.2          \\ \hline
		CLD     \cite{Wang_2021_CVPR}          & 54.2     & 42.0      & 80.7        & 31.5          \\ \hline
		HCSC    \cite{9880106}                 & 48.0     & 36.2      & 74.1        & 35.5          \\ \hline
		\textbf{IcicleGCN} & \pmb{80.7}     & \pmb{46.1}      & \pmb{95.5}        & \pmb{41.5}          \\
		\hline
\end{tabular}}
\end{table}

\begin{table}[!t]
\caption{The clustering performance w.r.t. \textbf{NMI(\%)} by different clustering methods on the benchmark datasets}\vskip 0.05 in
\label{table:NMI}
\centering
\setlength{\tabcolsep}{1.45mm}{
\begin{tabular}{|c|c|c|c|c|}
	\hline
	Dataset                                & CIFAR-10 & CIFAR-100 & ImageNet-10 & ImageNet-Dogs \\ \hline  \hline
	K-means \cite{macqueen1967some}        & 8.7      & 8.4       & 11.9        & 5.5           \\ \hline
	SC      \cite{zelnik2004self}          & 10.3     & 9.0       & 15.1        & 3.8           \\ \hline
	AC      \cite{gowda1978agglomerative}  & 10.5     & 9.8       & 13.8        & 3.7           \\ \hline
	NMF     \cite{2009Locality}            & 8.1      & 7.9       & 13.2        & 4.4           \\ \hline
	AE      \cite{bengio2006greedy}        & 23.9     & 10.0      & 21.0        & 10.4          \\ \hline
	DAE     \cite{vincent2010stacked}      & 25.1     & 11.1      & 20.6        & 10.4          \\ \hline
	DCGAN   \cite{radford2015unsupervised} & 26.5     & 12.0      & 22.5        & 12.1          \\ \hline
	DeCNN   \cite{zeiler2010deconvolutional} & 24.0     & 9.2       & 18.6        & 9.8           \\ \hline
	VAE     \cite{kingma2013auto}          & 24.5     & 10.8      & 19.3        & 10.7          \\ \hline
	JULE    \cite{yang2016joint}           & 19.2     & 10.3      & 17.5        & 5.4           \\ \hline
	DEC     \cite{xie2016unsupervised}     & 25.7     & 13.6      & 28.2        & 12.2          \\ \hline
	DAC     \cite{chang2017deep}           & 39.6     & 18.5      & 39.4        & 21.9          \\ \hline
	DDC     \cite{chang2019deep}           & 42.4     & -         & 43.3        & -             \\ \hline
	DCCM    \cite{wu2019deep}              & 49.6     & 28.5      & 60.8        & 32.1          \\ \hline
	IIC     \cite{ji2019invariant}         & 51.1     & 22.5      & -           & -             \\ \hline
	GATCluster \cite{niu2020gatcluster}    & 49.6     & 28.5      & 59.4        & 28.1          \\ \hline
	PICA    \cite{huang2020deep}           & 59.1     & 31.0      & 80.2        & 35.2          \\ \hline
	DRC     \cite{zhong2020deep}           & 62.1     & 35.6      & 83.0        & 38.4          \\ \hline
	CC      \cite{li2021contrastive}       & 70.5     & 43.1      & 86.2        & 40.1          \\ \hline
	CLD     \cite{Wang_2021_CVPR}          & 44.3     & 42.5      & 67.1        & 27.9          \\ \hline
	HCSC    \cite{9880106}                 & 40.7     & 36.1      & 64.7        & 35.5          \\ \hline
	\textbf{IcicleGCN} & \pmb{72.9}     & \pmb{45.9}      & \pmb{90.4}        & \pmb{45.8}          \\ \hline
\end{tabular}}
\end{table}

\begin{table}[!t] 
\caption{The clustering performance w.r.t. \textbf{ARI(\%)} by different clustering methods on the benchmark datasets}\vskip 0.05 in
\label{table:ARI}
\centering
\setlength{\tabcolsep}{1.45mm}{
\begin{tabular}{|c|c|c|c|c|}
\hline
Dataset                                & CIFAR-10 & CIFAR-100 & ImageNet-10 & ImageNet-Dogs \\ \hline  \hline
K-means \cite{macqueen1967some}        & 4.9      & 2.8       & 5.7         & 2.0           \\ \hline
SC      \cite{zelnik2004self}          & 8.5      & 2.2       & 7.6         & 1.3           \\ \hline
AC      \cite{gowda1978agglomerative}  & 6.5      & 3.4       & 6.7         & 2.1           \\ \hline
NMF     \cite{2009Locality}            & 3.4      & 2.6       & 6.5         & 1.6           \\ \hline
AE      \cite{bengio2006greedy}        & 16.9     & 4.8       & 15.2        & 7.3           \\ \hline
DAE     \cite{vincent2010stacked}      & 16.3     & 4.6       & 13.8        & 7.8           \\ \hline
DCGAN   \cite{radford2015unsupervised} & 17.6     & 4.5       & 15.7        & 7.8           \\ \hline
DeCNN   \cite{zeiler2010deconvolutional} & 17.4     & 3.8       & 14.2        & 7.3           \\ \hline
VAE     \cite{kingma2013auto}          & 16.7     & 4.0       & 16.8        & 7.9           \\ \hline
JULE    \cite{yang2016joint}           & 13.8     & 3.3       & 13.8        & 2.8           \\ \hline
DEC     \cite{xie2016unsupervised}     & 16.1     & 5.0       & 20.3        & 7.9           \\ \hline
DAC     \cite{chang2017deep}           & 30.6     & 8.8       & 30.2        & 11.1          \\ \hline
DDC     \cite{chang2019deep}           & 32.9     & -         & 34.5        & -             \\ \hline
DCCM    \cite{wu2019deep}              & 40.8     & 17.3      & 55.5        & 18.2          \\ \hline
IIC     \cite{ji2019invariant}         & 41.1     & 11.7      & -           & -             \\ \hline
GATCluster \cite{niu2020gatcluster}    & 40.8     & 17.3      & 55.2        & 16.3          \\ \hline
PICA    \cite{huang2020deep}           & 51.2     & 17.1      & 76.1         & 20.1          \\ \hline
DRC     \cite{zhong2020deep}           & 54.7     & 20.8      & 79.8        & 23.3          \\ \hline
CC      \cite{li2021contrastive}       & 63.7     & 26.6      & 82.5        & 22.5          \\ \hline
CLD     \cite{Wang_2021_CVPR}          & 31.9     & 26.4      & 62.6        & 14.1         \\ \hline
HCSC    \cite{9880106}                 & 29.5     & 29.5      & 55.9        & 20.9          \\ \hline
\textbf{IcicleGCN} & \pmb{66.0}     & \pmb{31.1}      & \pmb{90.5}        & \pmb{27.9}          \\  \hline
\end{tabular}}
\end{table}

\subsection{Comparisons with Other Clustering Methods}

In this section, we compare our IcicleGCN method with nineteen deep and non-deep clustering methods, including $K$-means\cite{macqueen1967some}, Spectral Clustering (SC) \cite{zelnik2004self}, Agglomerative Clustering (AC) \cite{gowda1978agglomerative}, Nonnegative Matrix Factorization (NMF) \cite{2009Locality}, Auto-Encoder (AE) \cite{bengio2006greedy}, Denoising Auto-Encoder (DAE) \cite{vincent2010stacked}, Deep Convolution Generative Adversarial Network (DCGAN) \cite{radford2015unsupervised}, Deconvolutional Network (DeCNN) \cite{zeiler2010deconvolutional}, Variational Auto-Encoder (VAE) \cite{kingma2013auto}, Jointly Unsupervised Learning (JULE) \cite{yang2016joint}, Deep Embedding Clustering (DEC) \cite{xie2016unsupervised}, Deep Adaptive image Clustering (DAC) \cite{chang2017deep}, Deep Discriminative Clustering (DDC) \cite{chang2019deep}, Deep Comprehensive Correlation Mining (DCCM) \cite{wu2019deep}, Invariant Information Clustering (IIC) \cite{ji2019invariant}, self-supervised Gaussian-ATtention network for image Clustering (GATCluster) \cite{niu2020gatcluster}, PartItion Confidence mAximisation (PICA) \cite{huang2020deep}, Deep Robust Clustering (DRC) \cite{zhong2020deep}, Contrastive Clustering (CC) \cite{li2021contrastive}, Cross-Level Discrimination (CLD) \cite{Wang_2021_CVPR} and Hierarchical Contrastive Selective
Coding (HCSC) \cite{9880106}. The clustering performances of different clustering methods w.r.t. ACC, NMI, and ARI are reported in Tables~\ref{table:ACC}, \ref{table:NMI}, and \ref{table:ARI}, respectively.

In terms of ACC, as shown in Table~\ref{table:NMI}, our IcicleGCN method outperforms or significantly outperforms the baseline clustering methods on all the four benchmark datasets. Especially, on the CIFAR-100, ImageNet-10, and ImageNet-Dogs datasets, our IcicleGCN method achieves ACC(\%) scores of 46.1, 95.5, and 41.5, respectively, while the best baseline method (i.e., CC) only obtains scores of 42.9, 89.5, and 34.2, respectively. In terms of NMI and ARI, similar advantages of IcicleGCN can also be observed . As shown in Table~\ref{table:NMI}, on the four benchmark datasets, our IcicleGCN method achieves NMI(\%) scores of 72.9, 45.9, 90.4, and 45.8, respectively, while the best baseline method only obtains scores of 70.5, 43.1, 86.2, and 40.1. As shown in Table~\ref{table:ARI}, IcicleGCN achieves ARI(\%) scores of 66.0, 31.1, 90.5, and 27.9 on the four datasets, while the best baseline method only obtains scores of 63.7, 26.6, 82.5, and 22.5, respectively. The experimental results in Tables~\ref{table:ACC}, \ref{table:NMI}, and \ref{table:ARI} confirm the superior clustering performance of the proposed IcicleGCN method over the baseline deep and non-deep clustering methods on the benchmark datasets.

\begin{table}[!t]
	\centering
	\caption{Influence of three losses in the ISM module and the JC-SLIM module}\vskip 0.05 in
	\label{table:ablationloss}
	\renewcommand\arraystretch{1.25} 
	\setlength{\tabcolsep}{2.6mm}{
		\begin{tabular}{cccc}   
			\toprule
			\multirow{2}{*}{Loss Function} & \multicolumn{3}{c}{ImageNet-Dogs} \\ \cline{2-4}
			& ACC       & NMI       & ARI       \\ \midrule
			$\mathcal{L}_{re-Loss}$ + $\mathcal{L}_{CIS-Loss}$ + $\mathcal{L}_{CCS-Loss}$                        & \pmb{41.5}      & \pmb{45.8}      & \pmb{27.9}      \\
			$\mathcal{L}_{re-Loss}$ + $\mathcal{L}_{CIS-Loss}$                                 & 25.8      & 23.2      & 12.5      \\
			$\mathcal{L}_{re-Loss}$ + $\mathcal{L}_{CCS-Loss}$                                   & 9.2       & 1.0       & 0.3       \\
			$\mathcal{L}_{re-Loss}$                                          & 6.9       & 0.4       & 0.0       \\
			\bottomrule
	\end{tabular}}
\end{table}

\begin{table}[!t]
	\caption{Influence of the GCNs and the auto-encoder (AE) in the M-GCN module}\vskip 0.05 in
	\label{table:ablationmodule}
	\centering
	\renewcommand\arraystretch{1.25} 
	\setlength{\tabcolsep}{2.45mm}{   
		\begin{tabular}{ccccc} 
			\toprule
			Dataset       & Ablation of Components     & ACC           &NMI          &ARI           \\ \midrule
			
			&\textit{with} GCNs and AE           & \pmb{41.5}  & \pmb{45.8}   & \pmb{27.9}     \\
			ImageNet-Dogs &\textit{without} GCNs        & 39.3      & 41.4   & 24.5     \\
			&\textit{without} GCNs and AE & 34.2   & 40.1   & 22.5     \\
			\bottomrule
		\end{tabular}\vskip 0.12 in
	}
\end{table}

\subsection{Ablation Study}
\label{sec:cmp_spectral}

In this section, we conduct ablation analysis to test the influences of different modules (and the components inside each module). We first test the clustering performance of IcicleGCN with different components in ISM and JC-SLIM or even the whole module removed in Section~\ref{sec:ablation_losses}. Then we test the clustering  performance with the components in M-GCN removed in Section~\ref{sec:ablation_mgcn}. Further, we test the influence of the multi-scale GCNs and the different neighborhood combinations in Sections~\ref{sec:multiway} and \ref{sec:neighbors}, respectively.

\subsubsection{Influences of the Components in ISM and JC-SLIM}
\label{sec:ablation_losses}

This section conducts ablation analysis on the components in ISM and JC-SLIM. Specifically, three losses are incorporated in the ISM and JC-SLIM modules, namely, the contrastive instance similarity loss $\mathcal{L}_{CIS-Loss}$, the contrastive cluster structure loss $\mathcal{L}_{CCS-Loss}$, and the instance reconstruction loss (of the auto-encoder)  $\mathcal{L}_{re-Loss}$, which jointly contribute to the self-supervised training of the CNN in our IcicleGCN framework. Notice that the auto-encoder serves as a bridge between the JC-SLIM module and the M-GCN module. Therefore, the reconstruction loss $\mathcal{L}_{re-Loss}$ will be preserved to keep this connection to M-GCN, while the other two losses will be removed and tested in this section. In fact, when the two contrastive losses are removed, the effects of the ISM module and the JC-SLIM module almost disappear, except that the auto-encoder still remains for the sake of M-GCN. As shown in Table~\ref{table:ablationloss}, both the contrastive instance similarity loss and the contrastive cluster structure loss play a substantial role in IcicleGCN. Especially, the joint incorporation of the three losses leads to the best clustering performance (w.r.t. ACC, NMI, and ARI) when compared to the variants with one or two components removed.

\subsubsection{Influence of the Components in M-GCN}
\label{sec:ablation_mgcn}

There are three streams of networks in the M-GCN module, including an auto-encoder and two streams of GCNs. In this section, we test the influence of the GCNs and the auto-encoder. Note that when the auto-encoder is removed, we will put back a nonlinear multi-layer MLP instead, so as to make the JC-SLIM module still functionable. As shown in Table~\ref{table:ablationmodule}, removing the GCNs degrades the ACC(\%), NMI(\%), and ARI(\%) scores from 41.5, 45.8, and 27.9 to 39.3, 41.4, and 24.5, respectively, while removing both the GCNs and the auto-encoder further degrades the ACC(\%), NMI(\%), and ARI(\%) scores to 34.2, 40.1, and 22.5, respectively, which demonstrate the contributions of the components in the M-GCN module. In the following, we will further test the influence of using different settings of the GCNs.

\begin{table}[!t]
	\centering
	\caption{The clustering performance of IcicleGCN using a single GCN and using multi-scale GCNs (with different neighborhood combinations)}\vskip 0.05 in
	\label{table:multiway}
	\renewcommand\arraystretch{1.25}
	\setlength{\tabcolsep}{2.8mm}{
		\begin{tabular}{cccc}    
			\toprule
			\multirow{2}{*}{Neighborhood Combinations in M-GCN} & \multicolumn{3}{c}{ImageNet-Dogs}                                                     \\ \cline{2-4}
			& ACC                      & NMI                      & ARI                      \\
			\midrule
			\emph{1}-NN                                        & 40.1                     & 43.5                     & 25.5                      \\
			\emph{3}-NN                                        & 39.8                     & 43.4                     & 25.6                      \\
			\emph{5}-NN                                        & 39.7                     & 43.4                     & 24.8                     \\
			\emph{10}-NN                                       & 39.6                     & 43.5                     & 25.3                      \\
			\midrule
			\emph{1}-NN+\emph{10}-NN                           & \textbf{41.5}                     & \textbf{45.8}                     & \textbf{27.9} \\
			\emph{3}-NN+\emph{10}-NN                           & 41.1                     & 45.3                     & 26.0 \\
			\emph{3}-NN+\emph{5}-NN                            & 41.4                    & 45.3                     & 27.0  \\
			\bottomrule
	\end{tabular}}
\end{table}

\subsubsection{Multi-scale GCNs  VS  Single GCN}
\label{sec:multiway}

{In this paper, the proposed IcicleGCN method is able to capture and propagate the multi-scale neighborhood information via two streams of GCNs, for which two $k$-NN graphs with different neighborhood sizes are constructed as the initial input. In this section, we test the clustering performance of IcicleGCN using the two-stream GCNs against using a single GCN. As shown in Table~\ref{table:multiway}, the incorporation of two scales of $k$-NN graphs (in the M-GCN module) yields consistently better clustering performance w.r.t. ACC, NMI, and ARI than using the GCN with a single-scale $k$-NN graph.}

\subsubsection{Influence of Different Neighborhood Combinations }
\label{sec:neighbors}

{In this section, we test the influence of using different combinations of $k$-NN graphs in the M-GCN module. In IcicleGCN, two streams of GCNs are utilized, each of which requires a $k$-NN graph as the input graph.} As shown in Table~\ref{table:multiway}, using two different scales of $k$-NN graphs is often beneficial for the clustering performance. {Particularly, using the combination of a $1$-NN graph and a $10$-NN graph leads to the optimal clustering performance, probably due to the complementariness between the nearest neighborhood and a relative larger neighborhood size. In this paper, we adopt the combination of a $1$-NN graph and a $10$-NN graph for our multi-scale neighborhood structure learning via the GCNs on all the benchmark datasets.}

\begin{figure}[!t]
	\begin{center}
		{\subfigure[CIFAR-10]
			{\includegraphics[width=0.229\linewidth]{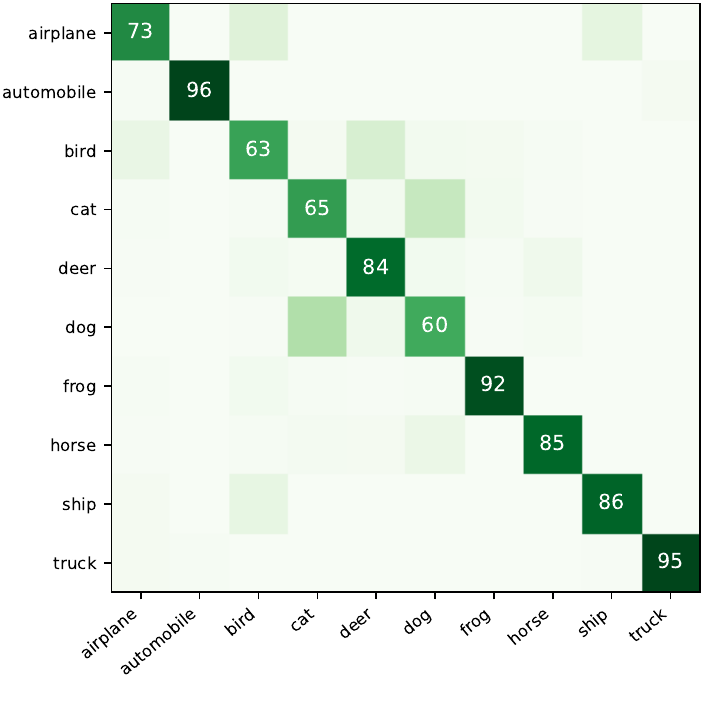}\label{fig:cifar10_matrix}}}
		{\subfigure[CIFAR-100]
			{\includegraphics[width=0.2584\linewidth]{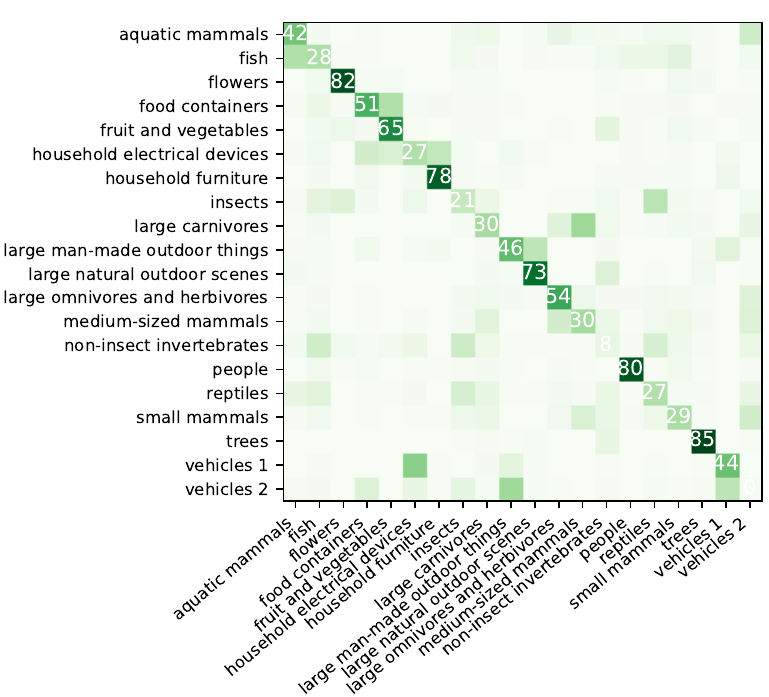}\label{fig:cifar100_matrix}}}
		{\subfigure[ImageNet-10]
			{\includegraphics[width=0.232\linewidth]{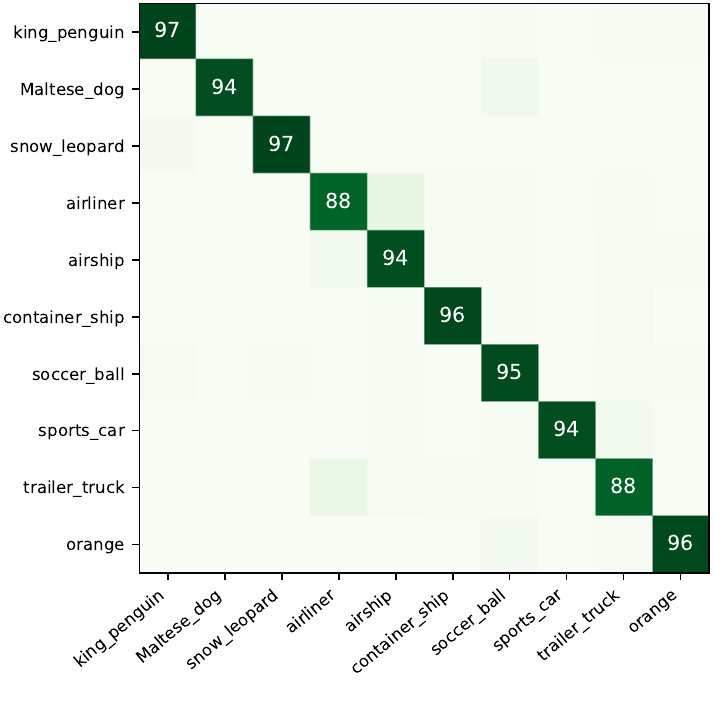}\label{fig:imageNet10_matrix}}}
		{\subfigure[ImageNet-Dogs]
			{\includegraphics[width=0.2495\linewidth]{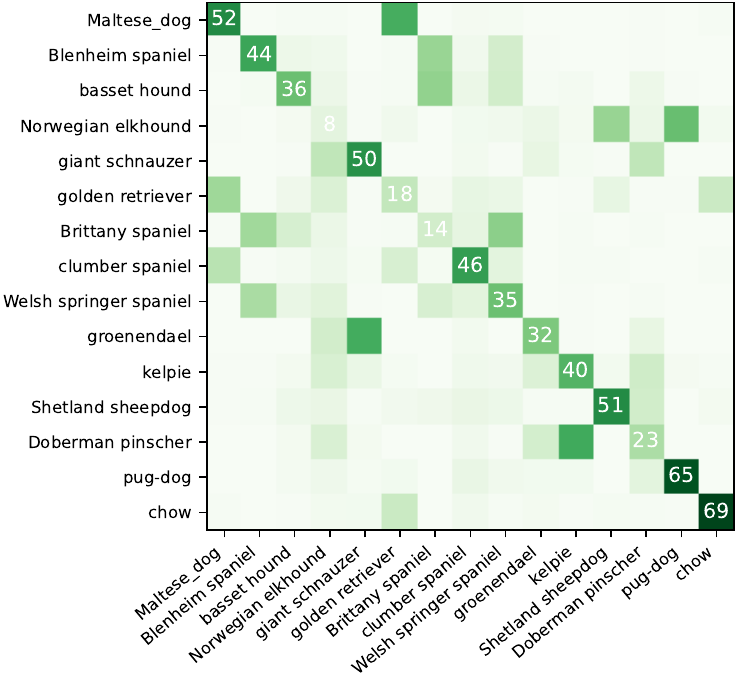}\label{fig:imageNet-dogs_matrix}}}
		\caption{Confusion matrices for the clustering results of IcicleGCN on the four image datasets.}
		\label{fig:confusion}
	\end{center}  
\end{figure}

\subsection{Visualization of Confusion Matrices}

In Fig.~\ref{fig:confusion}, we visualize the confusion matrices of the clustering results by IcicleGCN on the four datasets. The confusion matrices for CIFAR-10 and ImageNet-10 have a clear block diagonal structure that implies our IcicleGCN method successfully partitions most of the images into semantic clusters. For CIFAR-100 and ImageNet-Dogs, the block diagonal structure can still be observed, though it is not as clear as the other two confusion matrices, possibly due to the fact that (i) the images in CIFAR-100 are mostly small and blurry and that (ii) the categories of dogs in ImageNet-Dogs are sometimes difficult to be distinguished even for humans (as further illustrated in Fig.~\ref{fig:casestudy}). However, even on the challenging datasets of CIFAR-100 and ImageNet-Dogs, the proposed IcicleGCN method still exhibits a substantial advantage over the state-of-the-art deep clustering methods (as shown in Tables~\ref{table:ACC}, \ref{table:NMI}, and \ref{table:ARI}).

\subsection{Case Study}
\label{sec:}
		
\begin{figure}[!t] \vskip 0.15 in
	\begin{center}
		{\includegraphics[width=1\linewidth]{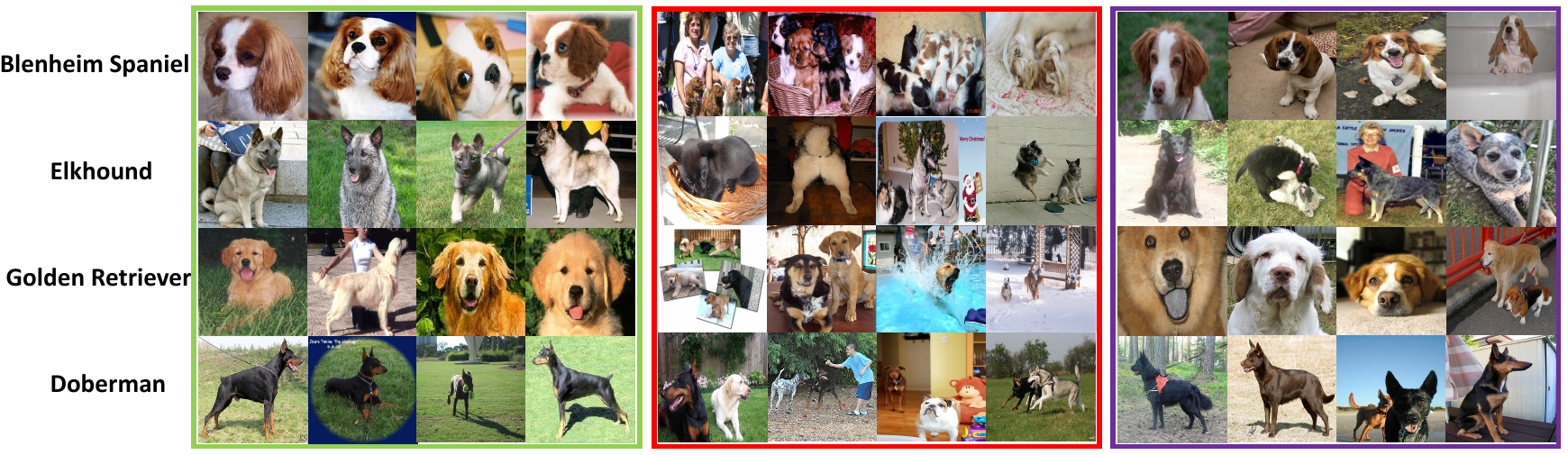}} 
		\caption{Case study on ImageNet-Dogs. The \textbf{\textit{left}} sub-figure shows the \textbf{successful} cases, the \textbf{\textit{middle}} sub-figure shows the \textbf{false negative} cases, and the \textbf{\textit{right}} sub-figure shows the \textbf{false positive} cases.}
		\label{fig:casestudy}
	\end{center}  \vskip 0.08 in
\end{figure}

To gain a deeper understanding of how our IcicleGCN method performs, we investigate the success and failure cases on the ImageNet-Dogs dataset. Specifically, we study three situations of four dog categories on ImageNet-Dogs, namely, (i) the \textit{successful} cases that correspond to the correctly clustered samples, (ii) the \textit{false negative} failure cases that correspond to the samples which belong to this cluster but are incorrectly assigned to other clusters, and (iii) the \textit{false positive} cases that correspond to the samples which do not belong to this cluster but are incorrectly assigned to this cluster. As shown in Fig.~\ref{fig:casestudy}, IcicleGCN successfully assigns many images of the same category to the same cluster. However, when there are multiple objects of different categories in the images, some of them may be assigned to the incorrect clusters (as shown in the middle sub-figure of Fig.~\ref{fig:casestudy}). Furthermore, some categories of dogs look very similar to each other, which may also contribute to the incorrect cluster assignments  (as shown in the right sub-figure of Fig.~\ref{fig:casestudy}). How to distinguish multiple objects in the same image and how to distinguish different fine-grained classes with similar appearance is still a very challenging problem for unsupervised learning and clustering.

\section{{Conclusion and Future Work}}
\label{sec:conclusion}
{In this paper, we propose a novel deep clustering approach called IcicleGCN, which bridges the gap between CNN and GCN as well as the gap between contrastive learning and multi-scale neighborhood structure learning for the deep image clustering task. The IcicleGCN approach consists of four main modules, i.e., the backbone, the ISM, the JC-SLIM, and the M-GCN.} With two augmented samples generated for each input image, the backbone with two weight-sharing views is utilized to extract their feature representations, which are then fed to the following three modules. In ISM and JC-SLIM, three types of losses are designed and jointly optimized, namely, the contrastive instance similarity loss, the contrastive cluster similarity loss, and the instance reconstruction loss (via an auto-encoder). {It is worth mentioning that the auto-encoder serves as a bridge between the JC-SLIM module and the M-GCN module.} In M-GCN, two streams of GCNs and the auto-encoder are iteratively and simultaneously updated with layer-wise representation fusion, upon which a joint self-adaptive learning mechanism is further incorporated to ensure the consistency of their last-layer clustering distributions. Experiments are conducted on four challenging image datasets, which have shown the advantageous clustering performance of IcicleGCN over the state-of-the-art deep clustering approaches.

{Note that this paper mainly focuses on the deep clustering task for image data. In future work, it may be a promising direction to extend the proposed deep clustering framework to other types of data, such as time series data \cite{zhong23_dtcc}. Besides, this paper performs the image-level clustering with each image treated as a sample, which may also be extended to the deep clustering at other levels of granularity, such as the pixel-level clustering (i.e., image segmentation \cite{ZHANG2021107885}) in the future research.}

\section*{Acknowledgments}
This work was supported by the NSFC (61976097, 62276277  \& U22A2095), and the Natural Science Foundation of Guangdong Province (2021A1515012203).

\bibliographystyle{elsarticle-num-names}
\bibliography{refs}

\newpage

\end{document}